\documentclass[10pt,twocolumn,letterpaper]{article}

\usepackage{iccv}              
\usepackage{graphicx}

\usepackage{multirow}
\usepackage{booktabs}
\usepackage{makecell}
\usepackage{graphicx}
\usepackage{colortbl}
\usepackage{xcolor}
\usepackage{adjustbox}

\definecolor{mygray}{gray}{0.9} 
\definecolor{lightgray}{gray}{0.85} 
\definecolor{iccvblue}{rgb}{0.21,0.49,0.74}
\usepackage[pagebackref,breaklinks,colorlinks,allcolors=iccvblue]{hyperref}

\def\paperID{4838} 
\def\confName{ICCV}
\def\confYear{2025}

\title{DynRsl-VLM: Enhancing Autonomous Driving Perception with Dynamic Resolution Vision-Language Models}

\author{Xirui Zhou\\
Xi’an Jiaotong University\\
Shaan Xi, China\\
{\tt\small ruixxove@stu.xjtu.edu.cn}
\and
Lianlei Shan\\
University of Chinese Academy of Sciences\\
Beijing, China\\
{\tt\small shanlianlei18@mails.ucas.edu.cn}
\and
Xiaolin Gui\\
Xi’an Jiaotong University\\
Shaan Xi, China\\
{\tt\small xlgui@mail.xjtu.edu.cn}
}

\begin{document}
\maketitle
\begin{abstract}
Visual Question Answering (VQA) models, which fall under the category of vision-language models, conventionally execute multiple downsampling processes on image inputs to strike a balance between computational efficiency and model performance. Although this approach aids in concentrating on salient features and diminishing computational burden, it incurs the loss of vital detailed information, a drawback that is particularly damaging in end-to-end autonomous driving scenarios. Downsampling can lead to an inadequate capture of distant or small objects such as pedestrians, road signs, or obstacles, all of which are crucial for safe navigation. This loss of features negatively impacts an autonomous driving system's capacity to accurately perceive the environment, potentially escalating the risk of accidents.
To tackle this problem, we put forward the Dynamic Resolution Vision Language Model (DynRsl-VLM). DynRsl-VLM incorporates a dynamic resolution image input processing approach that captures all entity feature information within an image while ensuring that the image input remains computationally tractable for the Vision Transformer (ViT). Moreover, we devise a novel image-text alignment module to replace the Q-Former, enabling simple and efficient alignment with text when dealing with dynamic resolution image inputs. Our method enhances the environmental perception capabilities of autonomous driving systems without overstepping computational constraints.
\end{abstract}
\section{Introduction}
Visual Question Answering (VQA) systems are emerging as a significant research focus in end-to-end autonomous driving. By integrating computer vision and natural language processing, VQA enables autonomous vehicles to comprehend visual environments and respond to queries in natural language. This synergy enhances environmental perception, real-time decision-making support, human-computer interaction, and anomaly detection with feedback, thereby deepening the system's understanding of road conditions and improving passenger experience~\cite{malla2023drama,ding2023hilm,xu2023drivegpt4,dewangan2023talk2bev}. Additionally, VQA enhances the interpretability and transparency of autonomous driving systems, contributing to their development, debugging, and compliance with regulatory requirements~\cite{atakishiyev2024explainableartificialintelligenceautonomous,atakishiyev2023explainingautonomousdrivingactions}. Looking ahead, with continuous technological advancements and industry collaboration, VQA holds vast potential in autonomous driving: technical maturity will steadily increase, multimodal integration will deepen, and system reliability will be continuously enhanced. 

Typically, Visual Question Answering (VQA) models, which are a subset of vision-language models, perform multiple downsampling operations on image inputs during processing. The primary purposes of this practice are to balance computational efficiency and model performance, while helping the model focus on key features, thereby preventing overfitting and reducing computational overhead~\cite{10.1007/978-3-031-31417-9_48,10.1007/978-3-031-64847-2_31}. The specific stages of image downsampling include not only initial image resizing but also the internal feature extraction processes within the model, such as convolutional layers and pooling operations; these are indispensable steps in the image processing pipeline designed to distill salient features from raw data~\cite{hesse2023contentadaptivedownsamplingconvolutionalneural}. However, in end-to-end autonomous driving  scenarios, downsampling can lead to feature loss, compromising environmental perception and increasing risks to driving safety. The reduced spatial resolution may fail to capture critical details of distant or small objects, such as pedestrians, road signs, or obstacles, which are essential for safe navigation. This limitation, in turn, adversely affects the autonomous driving system's comprehensive and accurate perception of the environment, which is fundamental for making correct predictions and executing reasonable planning maneuvers~\cite{ding2024holistic}. Consequently, in high-speed driving or complex traffic environments, the autonomous driving system may fail to detect and respond promptly to distant obstacles or traffic signals, thereby increasing the risk of accidents and compromising passenger safety.
To address the issue of detail feature loss caused by downsampling and to enable autonomous driving systems to fully perceive environmental information without missing any critical details—while ensuring that computational complexity remains within current technological capabilities and computational overhead is manageable—we developed the Dynamic Resolution Vision Language Model (DynRsl-VLM). Our contributions are as follows:
\begin{itemize}
    \item Dynamic Resolution Image Input Processing: We propose a dynamic resolution image input method that captures all entity feature information within an image while ensuring that the image input is not excessively large for the Vision Transformer (ViT) to handle effectively.
    \item Optimized DynRsl Image-Text Alignment Method: Based on the dynamic resolution image input, we designed a novel image-text alignment module to replace the Q-Former. This approach enables simple and efficient alignment with text when dealing with dynamic resolution image inputs.
\end{itemize}

\section{Related Works}
\subsection{Vision-Language Models}
Since the introduction of the Transformer architecture in 2017, vision-language models have advanced rapidly. The Transformer's powerful sequence modeling and global attention mechanisms revolutionized natural language processing and laid the groundwork for multimodal integration. Models like ViLBERT~\cite{lu2019vilbertpretrainingtaskagnosticvisiolinguistic} and LXMERT~\cite{tan-bansal-2019-lxmert} applied the Transformer to dual-stream processing of vision and language, separately encoding visual and textual information before fusing them, establishing the basis for multimodal learning. Subsequently, VisualBERT~\cite{li2019visualbertsimpleperformantbaseline} and VL-BERT~\cite{su2020vlbertpretraininggenericvisuallinguistic} further advanced the field by integrating visual and textual data within a unified Transformer architecture, enabling more effective multimodal representations.

Later, models like CLIP~\cite{radford2021learningtransferablevisualmodels} and ALIGN~\cite{jia2021scalingvisualvisionlanguagerepresentation} employed large-scale cross-modal contrastive learning to map images and texts into a shared embedding space, significantly improving performance in zero-shot and few-shot learning tasks. In multimodal dialogue and question answering, Visual Dialog~\cite{das2017visualdialog} enabled models to understand and generate image descriptions in multi-turn conversations. More recent work, such as Microsoft's VLMo~\cite{bao2022vlmounifiedvisionlanguagepretraining}, introduced a unified vision-language model capable of handling multiple tasks like visual question answering and image captioning.

Recent research emphasizes multimodal feature alignment, crucial for enhancing model performance. For instance, BLIP-2~\cite{li2023blip} introduces the Q-Former module—a lightweight query Transformer that extracts language-related information from visual features, achieving more efficient and fine-grained alignment. These advancements in pre-training and feature alignment have not only improved vision-language models across various tasks but also revitalized the development of multimodal artificial intelligence.
\subsection{VQA for End to End Driving}
In end-to-end autonomous driving, Visual Question Answering (VQA) is increasingly important for enhancing system understanding and responsiveness in complex environments. VQA has evolved from simple image understanding using template-based generative models and traditional convolutional neural networks (CNNs)~\cite{wu2016visualquestionansweringsurvey}, to models that combine CNNs with recurrent neural networks (RNNs) for better handling of visual and linguistic interactions~\cite{wu2016visualquestionansweringsurvey,VQA2015iccv,devlin2015languagemodelsimagecaptioning}.

Recent research optimizes VQA models specifically for autonomous driving contexts. Models like EM-VLM4AD~\cite{gopalkrishnan2024multi} and BEV-InMLLM~\cite{ding2024holistic} address the unique challenges of these environments. EM-VLM4AD is a lightweight, efficient vision-language model designed for question answering with multi-frame inputs, reducing computational overhead through architectural optimization. BEV-InMLLM combines Bird's Eye View (BEV) representations with other sensor data like images and LiDAR to achieve a comprehensive understanding of the environment, better capturing spatial relationships and contextual information. These advancements enhance autonomous driving systems' ability to understand and respond accurately to scene-related questions in complex, dynamic environments, laying a solid foundation for future developments in autonomous driving technology.

\begin{figure*}[htbp]
  \centering
  \includegraphics[width=0.9\textwidth]{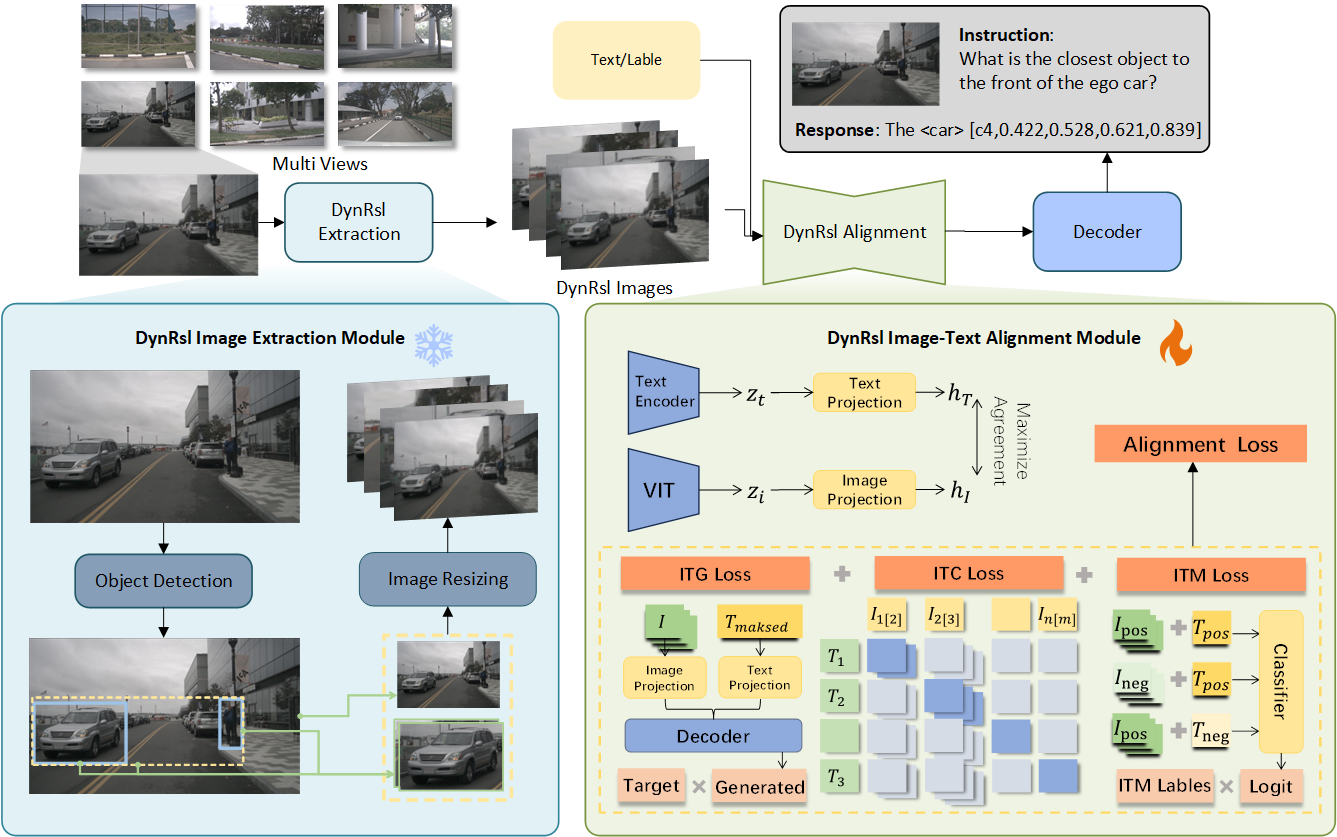}
  \caption{\textbf{The architecture of our model that acquires multi-resolution images, performs visual-text alignment, and conducts efficient computations.}}
  \label{fig:model_overview}
\end{figure*}

\subsection{Fine-grained contrastive learning}
Fine-grained contrastive learning (FGCL) goes beyond standard contrastive methods by emphasizing subtle, localized differences within images. While BLIP-2 largely employs a Q-Former for higher-level vision-language alignment~\cite{li2023blip2bootstrappinglanguageimagepretraining}, our approach refines this strategy by introducing patch-to-patch contrast, directly matching each image patch with its relevant textual segment. This local alignment better captures nuanced features such as object parts, small signs, or intricate patterns.

FGCL has demonstrated effectiveness in fine-grained classification, small-object detection, and detailed multi-modal scenarios. Instead of merely treating each image or text instance as a single global feature, patch-level contrasts encourage the model to focus on every meaningful region. Recent studies show that such region-level or saliency-based augmentations yield representations more sensitive to minor distinctions, enhancing both recognition and retrieval tasks.

When extended to vision-language modeling, FGCL’s patch-to-text alignment strengthens cross-modal representations. By contrasting corresponding patches and words or phrases, the model learns a more detailed mapping of visual concepts to linguistic descriptions. However, its implementation can be more complex than global contrastive approaches: additional steps are needed to identify appropriate local patches, and carefully chosen negatives are crucial to avoid confusion among highly similar regions.

In our research, dynamic resolution inputs allow the model to retain essential fine details, and our patch-to-patch FGCL fully exploits these details for precise cross-modal understanding. This synergy is invaluable in autonomous driving, where identifying distant or small objects—pedestrians, signs, or hazards—can be critical for safe navigation.

\subsection{Dynamic Resolution Processing}
Dynamic Resolution Processing has seen tremendous development in recent years.
ControlCap~\cite{zhao2025controlcap} introduces a controllable region-level captioning method, emphasizing the importance of regional information in enhancing the accuracy of scene understanding. Meanwhile, DynRefer~\cite{zhao2024dynrefer} optimizes information extraction for multimodal tasks by dynamically adjusting the resolution of different regions within an image.
Meanwhile, in the context of autonomous driving with Vision-Language Models (VLMs), precise feature extraction and dynamic resolution adjustment can significantly enhance the recognition of key driving information while improving the computational efficiency of the system, thereby enhancing overall driving safety and performance.
\section{Proposed Method}
\subsection{Overview}
DynRsl-VLM consists of a dynamic-resolution(DynRsl) image extraction module, a customized DynRsl image-text alignment module, and the pre-trained Transformer Decoder. The dynamic resolution image extraction module generates multi-resolution image inputs through global object detection and Regions of Interest (ROI) merging operations, effectively preserving image details while ensuring computational efficiency. The image embedding module uses a frozen Vision Transformer (ViT) to extract image embeddings from dynamic resolution images taken from six different perspectives. Finally, the extracted image and text embeddings are aligned through an customized alignment module, with the resulting features input into the Transformer Decoder for visual-language training. The model architecture is shown in Figure \ref{fig:model_overview}.

\begin{figure}[htbp]
  \centering
  \includegraphics[width=\linewidth]{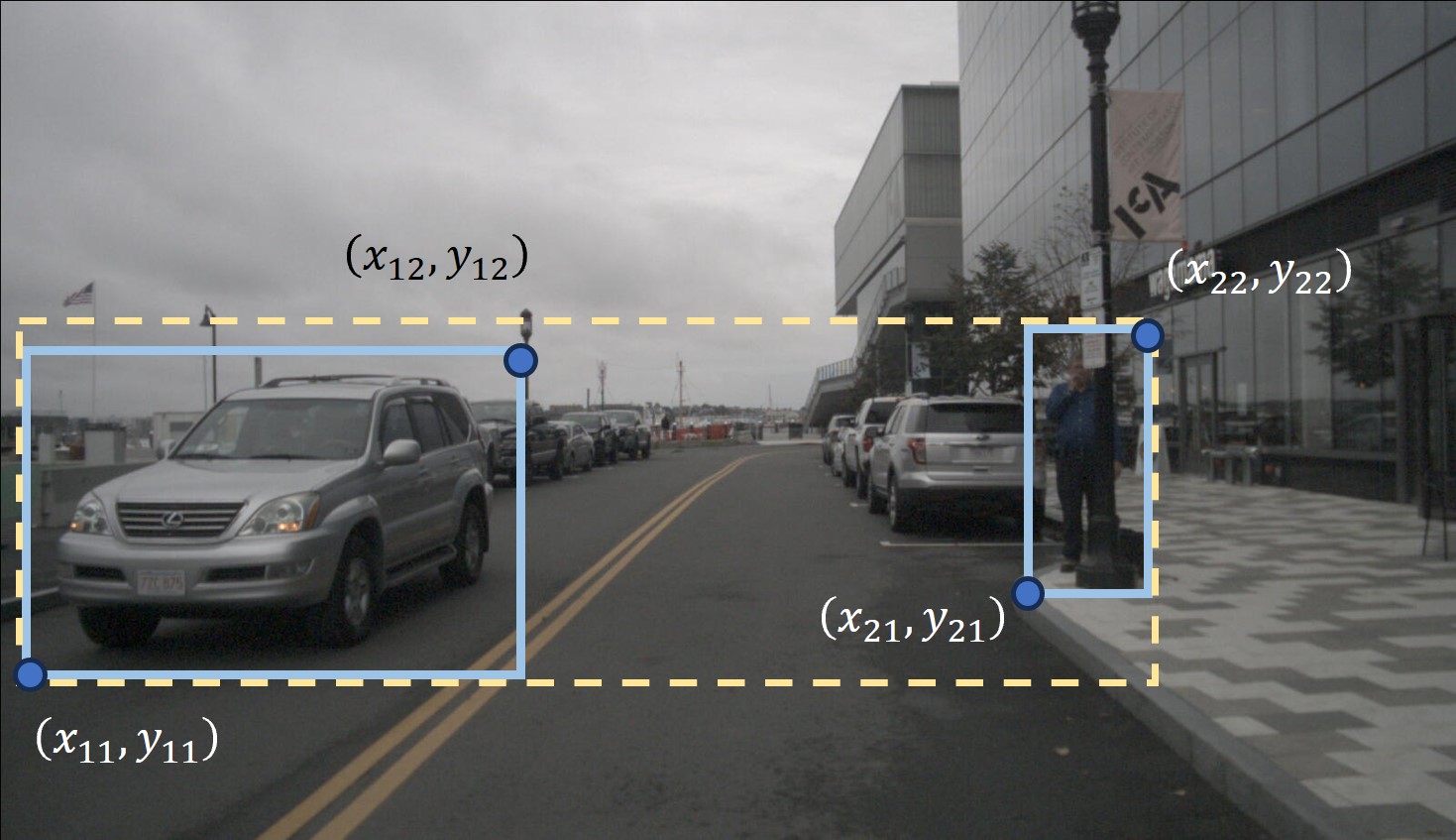}
  \caption{\textbf{Method for obtaining Region Images.} This diagram illustrates the approach for acquiring Region Images, which include both individual entity regions and combined regions. The blue solid-line boxes represent the ROIs, while the yellow dashed-line boxes denote the combined region.}
  \label{fig:dynrsl}
\end{figure}

\subsection{DynRsl Image Extraction}
The reduction in driving safety details due to low resolution and downsampling can result in the loss of features of smaller entities in the image, such as distant vehicles or pedestrians. This feature loss hinders the VQA model from obtaining comprehensive information, leading to flawed responses and subsequent safety risks.

To preserve these image details without incurring excessive computational costs, we utilize a DynRsl image extraction module to process high-resolution original images. The high-resolution image initially undergoes global object detection (using YOLOv8\cite{10533619} in our case) to generate detection results. All vehicles and pedestrians in the high-resolution image are labeled efficiently, preserving the critical details necessary for further analysis. These regions containing vehicles and pedestrians are then cropped from the original image at high resolution and referred to as Regions of Interest (ROIs), thereby preserving all essential image details.

However, simply listing all regions of interest (ROIs) is inadequate for extracting comprehensive information from an image, as entities within the image interact with one another. For instance, the behavior of a pedestrian standing at an intersection is influenced by vehicles crossing that intersection. Therefore, considering the potential relationships between different entities is crucial. To explore these relationships, we perform a merging operation on the ROIs, as illustrated in Figure \ref{fig:dynrsl}. First, we compute all possible combinations of ROIs (each combination containing two or more ROIs), and then we consolidate each combination into a combined region. The bounding box of the combined region is determined by the constituent ROIs. The specific method is as follows:

\begin{figure}[htbp]
  \centering
  \includegraphics[width=\linewidth]{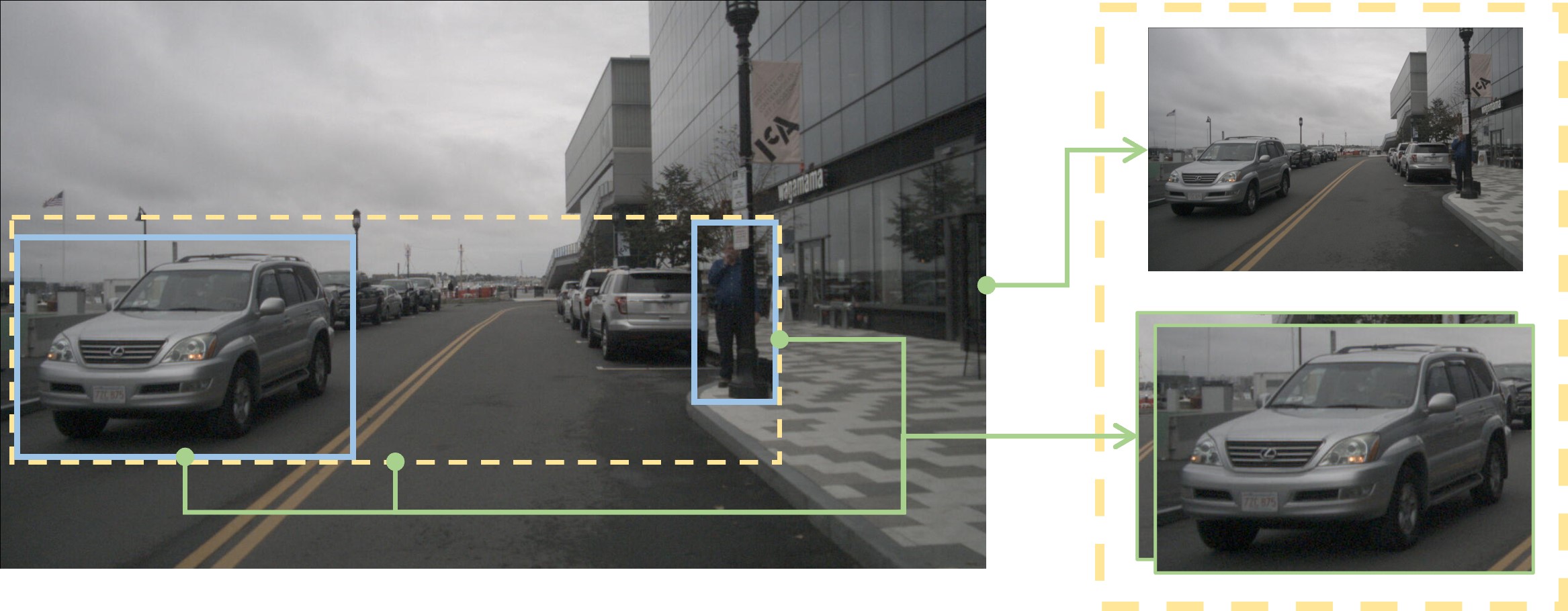}
  \caption{\textbf{Method for obtaining DynRsl image inputs.} This diagram shows how DynRsl image inputs are derived from the original high-resolution image, where the overall image resolution is reduced while maintaining the clarity of Region Images.}
  \label{fig:image_input}
\end{figure}

As shown in Figure \ref{fig:dynrsl}, the ROIs that constitute the Combined Region can be represented by a set of bounding boxes \( B = \{b_1, b_2, \ldots, b_n\} \), where each bounding box \( b_i \) is defined by its coordinates \((x_{i1}, y_{i1}, x_{i2}, y_{i2})\), corresponding to the bottom-left corner \((x_{i1}, y_{i1})\) and the top-right corner \((x_{i2}, y_{i2})\). The coordinates of the merged bounding box, or the bounding box of the Combined Region \( C \), are computed as follows:
\begin{equation}
\label{eq:merged_bbox}
\begin{aligned}
x_{\text{min}} &= \min\limits_{i=1}^{n} x_{i1}, &\quad
y_{\text{min}} &= \min\limits_{i=1}^{n} y_{i1}, \\
x_{\text{max}} &= \max\limits_{i=1}^{n} x_{i2}, &\quad
y_{\text{max}} &= \max\limits_{i=1}^{n} y_{i2}.
\end{aligned}
\end{equation}

Therefore, the coordinates of the combined bounding box \( C \) are:\((x_{\text{min}}, y_{\text{min}}, x_{\text{max}}, y_{\text{max}})\). Finally, we crop the combined regions from the original image at its original resolution based on these bounding boxes and refer to them, along with the ROIs, as Region Images. Through this process, our Region Images encompass not only the Regions of Interest (ROIs) that capture all detailed features but also the combined regions that illustrate the relationships between entities. As depicted in Figure \ref{fig:image_input}, When integrated with the downsampled original image representing global features, we obtain an image input rich in details without incurring excessive computational cost. For any given entity in the image, we retain its detailed features through the ROI, comprehend the spatial relationships between entities via the combined regions, and acquire global information from the low-resolution overall image, thereby achieving dynamic resolution in the image input.

\subsection{Alignment}
This section outlines our approach for aligning dynamic resolution visual inputs with their corresponding textual inputs, a process that is crucial for visual-language models. The principles of the alignment module are illustrated in the figure below.

We leverage the Vision Transformer (ViT) architecture to extract rich, context-aware features from each resolution image. It is important to note that the ViT is frozen during the pretraining phase to preserve the learned features. Subsequently, we enable attention mechanisms between the image features at different resolutions, which facilitates the exploration of spatial relationships between entities, as well as between parts and the whole.

\begin{figure*}[htbp]
  \centering
  \includegraphics[width=0.8\textwidth]{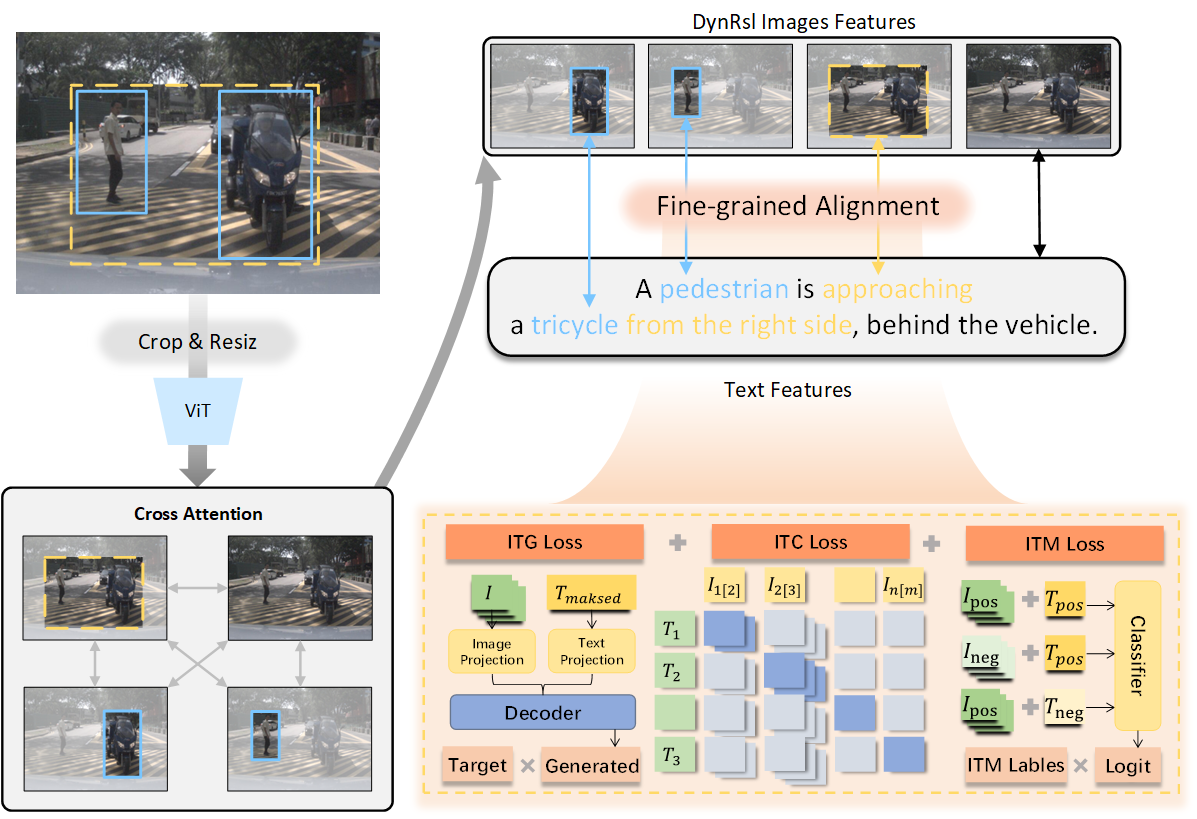}
  \caption{\textbf{Architecture of the alignment module and the losses employed during model training.} First, we extract features from the image input using the frozen ViT, and then apply attention between the image features at different resolutions. The resulting features are considered our dynrsl image features, which are aligned with the text features. This alignment involves matching multiple image features to a single text feature. This alignment method offers finer granularity and is achieved through the design of three distinct loss functions.}
  \label{fig:align_loss}
\end{figure*}

This approach leverages the Transformer's self-attention mechanism to capture both local and global dependencies, enabling the extraction of rich, context-aware features. Using dynamic resolution inputs enhances the model's ability to understand complex driving environments while staying within the ViT's processing capacity, preventing computational overload. The resulting features are referred to as DynRsl image features. In the subsequent alignment process, these features are paired with text features. Unlike previous works with one-to-one pairing, we pair a text feature with multiple DynRsl image features of varying resolutions, as illustrated in Figure~\ref{fig:align_loss}. This one-to-many pairing enables the model to align textual descriptions with visual features at different levels of detail, enhancing the richness and precision of the alignment. This is particularly advantageous in our tasks, allowing for efficient and detailed environmental perception essential for autonomous driving.

For the language input, we use a pre-trained language model like BERT~\cite{devlin2019bertpretrainingdeepbidirectional} to encode the descriptions into contextualized feature vectors \(z_T\) .

To reconcile differences in dimensionality and representation between visual and textual embeddings, we introduce two projection heads, as shown in Figure \ref{fig:align_loss}. The Image Projection Head \(P_I\) and the Text Projection Head \(P_T\) transform the visual and textual embeddings \(z_I\) and \(z_T\) into a common feature space: \(h_I = P_I(z_I), \quad h_T = P_T(z_T)\). These projection heads employ ReLU as the activation function to introduce non-linearity.

To validate the effectiveness of our feature alignment module, we integrate the projection heads into the BLIP-2~\cite{li2023blip} model, replacing the Q-Former. We conduct the same pre-training tasks as BLIP-2, connecting the projection heads to frozen image and text encoders and using image-text pairs for pre-training. We jointly optimize three pre-training objectives to enhance feature alignment.

\noindent\textbf{Image-Text Contrastive Learning (ITC)} aims to align image representations and text representations to maximize the mutual information between them. This is achieved by comparing the similarity of positive image-text pairs with that of negative pairs. We align the projected features of the DynRsl image and text, namely \(h_I\) and \(h_Tcls\), where \(h_Tcls\) represents the output embedding of the [CLS] token. Since \(h_I\) contains multiple output embeddings (one for each patch), we first compute the pairwise similarity between each patch output and \(h_Tcls\), then select the highest similarity as the image-text similarity. Because two different projection heads are used, there is no concern about information leakage between the image features and text features. Similar to BLIP2~\cite{li2023blip}, we use intra-batch negative samples instead of the momentum queue used in BLIP.

In contrastive learning, both Alignment and Uniformity are crucial. Alignment ensures that corresponding features (positive examples) are as close as possible after mapping, while Uniformity ensures that mapped features retain distinctive information and are uniformly distributed in the embedding space.

For example, if we have images of a cat and a dog, along with the text descriptions "cat" and "dog," Alignment means the cat image's features should closely match the "cat" text's features. Uniformity means the features of the cat and dog images should be uniformly distributed, not clustered together. If all features converge to a single point, causing different features to collapse to the same constant after projection, this indicates that the distinctive features of cats and dogs have been lost—a situation referred to as model collapse.

We employ a symmetric InfoNCE loss function as our contrastive learning objective to effectively align DynRsl image features and textual features. The symmetric loss considers both the image-to-text and text-to-image directions, enhancing the model's alignment performance.

The similarity between the DynRsl image features \(h_I\) and the textual feature \(h_T\) is computed using cosine similarity:
\begin{equation}
\text{$sim$}(h_I, h_T) = \frac{h_I \cdot h_T}{\|h_I\| \|h_T\|},
\end{equation}
where \(h_I\) and \(h_T\) are the normalized features vectors of the image and text, respectively; \(\cdot\) denotes the dot product, and $\| \cdot \|$ represents the Euclidean norm. The symmetric InfoNCE loss is defined as:
\begin{equation}
    L_{\text{$Sym$}} = \frac{1}{2} \left( L_{\text{$InfoNCE$}}^{\text{$i2t$}} + L_{\text{$InfoNCE$}}^{\text{$t2i$}} \right).
\end{equation}
The symmetric InfoNCE loss \(L_{\text{$Sym$}}\) is the average of the losses from both image-to-text and text-to-image directions. The image-to-text InfoNCE loss \(L_{\text{$InfoNCE$}}^{\text{$i2t$}}\) measures the alignment of image features $h_{I_i}$ with their corresponding text features $h_{T_i}$ relative to other texts in the batch. The text-to-image InfoNCE loss $L_{\text{$InfoNCE$}}^{\text{$t2i$}}$ measures the alignment of text features $h_{T_i}$ with their corresponding image features $h_{I_i}$, similarly to \(L_{\text{$InfoNCE$}}^{\text{$i2t$}}\). The single-directional InfoNCE losses are formulated as:
Image-to-text direction:
\begin{equation}
    L_{\text{$InfoNCE$}}^{\text{$i2t$}} = -\frac{1}{N} \sum_{i=1}^{N} \log \left( \frac{\exp\left( \frac{\text{$sim$}(h_{I_i}, h_{T_i})}{\tau} \right)}{\sum_{j=1}^{N} \exp\left( \frac{\text{$sim$}(h_{I_i}, h_{T_j})}{\tau} \right)} \right) .
\end{equation}
Text-to-image direction:
\begin{equation} 
    L_{\text{$InfoNCE$}}^{\text{$t2i$}} = -\frac{1}{N} \sum_{i=1}^{N} \log \left( \frac{\exp\left( \frac{\text{$sim$}(h_{T_i}, h_{I_i})}{\tau} \right)}{\sum_{j=1}^{N} \exp\left( \frac{\text{$sim$}(h_{T_i}, h_{I_j})}{\tau} \right)} \right) .
\end{equation}
In these equations, \(N\)is the batch size, and \(\tau\) is a temperature hyperparameter controlling the sharpness of the distribution, where $h_{I_i}$ and $h_{T_i}$ denote the projected features of the \(i-th\) DynRsl image and text in the batch, respectively.

The symmetric InfoNCE loss encourages the model to align matching image-text pairs while pushing apart non-matching pairs by considering both embedding mapping directions. This results in a more robust and effective alignment between the visual and textual modalities.

\noindent\textbf{Image-Guided Text Generation (ITG)} We employ the Image-Guided Text Generation (ITG) loss to train the projection heads for extracting features beneficial to text generation, conditioned on the input image. Unlike Q-Former~\cite{li2023blip}, which uses self-attention masks to control the interaction between queries and text, we directly pass the image embeddings—obtained from dynamic resolution image inputs—and masked text tokens through the projection heads to obtain features. These features are then input together into the same decoder used by Q-Former for prediction.  In this setup, a single text embedding is paired with multiple image embeddings corresponding to different resolutions, forming a one-to-many relationship, as illustrated in Figure~\ref{fig:align_loss}. Although this approach does not allow interaction between the image and text during the feature extraction phase, it simplifies the process and improves efficiency while still aiding the projection heads in extracting features suitable for text generation. We also replace the [CLS] token with a new [DEC] token, making it the first text token to indicate the decoding task.

\noindent\textbf{Image-Text Matching (ITM)} aims to learn fine-grained alignment between image and text representations. It is a binary classification task where the model predicts whether an image-text pair is a positive sample (matched) or a negative sample (mismatched). In our approach, we input the features of each image patch along with the text features into a binary linear classifier to obtain a logit for each patch-text pair. We then average these logits to produce the overall matching score between the image and text. This method allows us to perform multimodal feature alignment at a finer granularity compared to Q-Former~\cite{li2023blip}.

Additionally, we adopt the hard negative sampling strategy proposed by Li et al.~\cite{li2022blip} to create informative negative sample pairs. Negative samples are selected based on the image-text similarities computed during the ITC process; specifically, we randomly choose negative pairs with higher similarity scores to form more challenging negatives.

\section{Experiments}
\begin{table*}
    \centering
    \label{tab:sargd_ablation_study}
    \begin{adjustbox}{max width=\textwidth}
    \begin{tabular}{l|c c c c c c | c c | c c c c c c | c}
    \Xhline{1.5pt}
    \rowcolor{mygray}
    &\multicolumn{6}{c|}{\it \textbf{Perception}}
    &\multicolumn{2}{c|}{\it \textbf{Prediction}}
    &\multicolumn{6}{c|}{\it \textbf{Risk}}
    &\\

    \rowcolor{mygray}

    \multirow{-2}{*}{\textbf{Method}} &  Dis$\downarrow$ &  Sped $\downarrow$ & \# Ins $\downarrow$  & Clos $\uparrow$   & Sta $\uparrow$ & SameR $\uparrow$
    & Mot $\downarrow$ & Sta $\uparrow$
    & App $\uparrow$ & LaneC $\uparrow$ & Onco $\uparrow$&  Cro $\uparrow$  & Over $\uparrow$  & Brak $\uparrow$  &  \multirow{-2}{*}{\thead{\it \textbf{Planning} \\ \textbf{with  Reasoning} } $\uparrow$ }

    \\
  
    \hline \hline
    BLIP-2$^*$ &29.3 & 5.6 & 4.4 & 18.5 & 15.9 & 23.8 & 8.7 & 38.7 & 6.1 & 10.6 & 15.4 & 16.7 & 3.7 & 21.5 & 24.9\\
    
    MV-MLLM  & 26.8 & 5.3 & 3.9 & 28.2 & 17.7 & 31.5 & 6.8 & 43.6 & 15.2 & 19.3  & 18.4 & 22.7& 9.6 & 22.7 & 30.1 \\
    
    {BEV-InMLLM}  & \textbf{23.3} & \textbf{3.6} & \textbf{3.2} & \textbf{33.6} & \textbf{18.9} & \textbf{31.6} & \textbf{3.8} & \textbf{45.2} &  \textbf{16.8} & \textbf{21.0}  & \textbf{19.7} & \textbf{23.9} &\textbf{10.5} & \textbf{27.5} & \textbf{33.3}\\
    \midrule
    MiniGPT-4$^*$ & 30.2 & 6.2 & 6.3 & 20.2	&17.3 & 24.2 &	8.7 & 39.6 & 7.8 & 12.5 & 16.9 & 18.7& 4.8 & 21.8 &26.3\\
    
    MV-MLLM  & 28.6 & 4.7 & 4.1 & 27.5 & 18.5 & 30.7 & 7.2 & 44.2 & 15.5 & 18.9  & 19.1 & 23.3& 8.2 & 23.1 & 32.3 \\
    
    BEV-InMLLM & \textbf{23.6} & \textbf{3.1} & \textbf{3.8} & \textbf{32.9} & \textbf{19.2} & \textbf{31.5} & \textbf{4.2} & \textbf{46.5} &  \textbf{17.3} & \textbf{20.5}                & \textbf{21.5} & \textbf{24.5}& \textbf{9.4} & \textbf{26.8} & \textbf{35.6}\\
    \midrule
    Video-LLama & 29.9 & 6.5 & 5.4 & 22.3 & 16.7 & 20.9 & 9.3 & 39.3 & 6.2 & 10.9 & 16.2 & 18.4 & 4.1 & 21.3 & 25.3\\              
    MV-MLLM & 28.9 & 6.2 & 2.4 & 27.9 & 19.6 & 30.9 & 9.3 & 44.3 & 16.5 & 18.7 & 19.9 & 23.0 & 6.5 & 26.6 & 31.4\\
    
    {BEV-InMLLM} & 24.5 & 3.5 & 4.2 & 31.6 & 19.0 & 34.6 & 4.1 & 44.7 & 17.7 & 22.5 & 21.4 & 26.1 & 8.7 & 27.9 & 35.2\\
    {Ours} & \textbf{23.9} & \textbf{3.4} & \textbf{4.0} & \textbf{31.9} & \textbf{19.3} & \textbf{34.8} & \textbf{4.0} & \textbf{44.9} & \textbf{17.9} & \textbf{22.7} & \textbf{21.5} & \textbf{26.4} & \textbf{9.0} & \textbf{28.2} & \textbf{35.5}\\
    \bottomrule
    \end{tabular}
    \end{adjustbox}
     \vspace{-3.0mm}
     \caption{\textbf{Comparison of experimental results between our model and baseline models across various tasks.} This table presents the performance of our proposed model (Ours) in comparison with baseline models—BLIP-2, MV-MLLM, and BEV-InMLLM—on perception, prediction, risk assessment, and planning tasks. Our model achieves improved results across all tasks, demonstrating its effectiveness in autonomous driving scenarios.
    }
    \label{tab:sota}
\vspace{-1.0em}
\end{table*}
\subsection{Datasets}
We evaluate our model on the NuInstruct~\cite{ding2024holistic} dataset, which was constructed to facilitate research in visual question answering for autonomous driving. This dataset is derived from the NuScenes dataset~\cite{caesar2020nuscenes} and consists of 11,850 keyframes sampled from 850 videos. After filtering, it contains 55,204 unique instances that appear a total of 295,828 times across all keyframes, averaging approximately 24.96 instances per keyframe.

Using an SQL-based method, the original authors generated 91,355 instruction-response pairs encompassing four primary tasks: Perception, Prediction, Risk Assessment, and Planning with Reasoning. These tasks are further divided into 17 sub-tasks, providing a comprehensive set of challenges for evaluating autonomous driving systems.

One of the notable features of NuInstruct~\cite{ding2024holistic} is its multi-view information coverage, which sets it apart from other single-view benchmarks. Statistical analysis shows that to answer the instructions effectively, the system needs to consider multiple perspectives rather than relying on a single view. Specifically:

\begin{itemize}
    \item Balanced Multi-View Information: In the Perception and Prediction tasks, the distribution across various views is relatively even, indicating that the data generation method effectively produces balanced multi-view information.
    \item Front-Focused Responses: For Reasoning and Risk Assessment tasks, responses predominantly rely on information from the front, left-front, and right-front views. This aligns with real-world driving behavior, where drivers primarily focus on the areas ahead and to the sides to make decisions.
\end{itemize}

\noindent\textbf{Data Split.} Following the strategy used in the original NuInstruct paper~\cite{ding2024holistic}, we split the dataset into training, validation, and testing sets with a ratio of 7.5:1.5:1.5, respectively. Specifically, the 850 videos from NuScenes~\cite{caesar2020nuscenes} are divided accordingly, ensuring that the distribution of tasks and views remains consistent across splits. We train our models on the training set and select the best-performing model on the validation set for evaluation on the test set.

By utilizing NuInstruct and adhering to the same data splitting strategy, we ensure the comparability and validity of our experimental results.

\subsection{Implementation Details}
We evaluate our Dynamic Resolution Vision Language Model (DynRsl-VLM) on three base multimodal large language models (MLLMs): BLIP-2~\cite{li2023blip}, Video-LLaMA~\cite{zhang2023video}, and MiniGPT-4~\cite{zhu2023minigpt}. To adapt BLIP-2 and MiniGPT-4—originally designed for image inputs—to handle video inputs, we incorporate a spatiotemporal adapter (ST-Adapter)~\cite{pan2022parameter} following the method proposed in \cite{ding2023hilm}, while keeping their pre-trained parameters unchanged.

All MLLMs are initialized with their official pre-trained weights, and we freeze these weights during training. We only train the parameters of the ST-Adapters and our additional modules, which include the projection heads and the alignment module, which is a modification of Q-Former.

For the visual feature dimension $D_{\text{vis}}$ and the response feature dimension $D_{\text{resp}}$, we set both to 1,408, matching the hidden feature dimension used in BLIP-2. We set specific parameters such as the number of views $N_{\text{view}} = 6$ consistent with the multi-view nature of the NuInstruct dataset.

In our experiments, the input images are resized and cropped to a spatial size of 224 × 224 pixels. For video inputs, each video is uniformly sampled to select 3 frames. This approach allows us to capture temporal information while keeping computational requirements manageable.

We employ the AdamW optimizer~\cite{loshchilov2017decoupled} with an initial learning rate of $1 \times 10^{-4}$. The learning rate is scheduled using a cosine annealing scheduler~\cite{loshchilov2016sgdr}. All models are trained for 20 epochs. During training, we preserve the pre-trained weights of the base MLLMs and focus on optimizing our added components to ensure efficient convergence and to leverage the rich representations learned by the foundational models.

\subsection{Evaluation Metric}
The NuInstruct dataset includes a variety of tasks, making it challenging to evaluate all tasks with a single metric. Therefore, we employ a combination of metrics tailored to different task types, consistent with those used in the original NuInstruct paper, as summarize in Table~\ref{tab:metrics}. 

By employing the same evaluation metrics and methodologies as the original NuInstruct paper, our results are directly comparable to existing benchmarks. This consistency enhances the validity of our performance assessments and facilitates fair comparisons with other models.

\begin{table}
    \centering
    \begin{adjustbox}{max width=\linewidth}
    \begin{tabular}{l|c|c}
    \Xhline{1.5pt}
     \rowcolor{mygray}{\it \small \textbf{Task}}
    &{\small  \it \textbf{SubTask}}
    &{\small  \it \textbf{Metrics}}
    \\
    \midrule
    \multirow{2}{*}{Perception} &  Distance, Speeds, Instance Number & {MAE $\downarrow$}\\
    & Closest, Status, Same Road & Accuracy $\uparrow$\\
    \midrule
     \multirow{2}{*}{Prediction} &  Motion Ego, Motion Others & {MAE $\downarrow$}\\
     & Status Ego, Status Others & {Accuracy $\uparrow$}\\
    \midrule
   Risk & All &  MAP $\uparrow$\\
        \midrule
        Reasoning & All & BLEU~\cite{papineni2002bleu} $\uparrow$\\
    \bottomrule
    \end{tabular}
        \end{adjustbox}
        \vspace{-3.0mm}
         \caption{ \textbf{Evaluation metrics for different tasks.} $\downarrow$ represents the lower the scores, the better the results, while $\uparrow$ means the higher the scores, the better the results. `MAE' indicates the mean absolute error. `All' means the all subtasks.
    }
    \label{tab:metrics}
\end{table}

\subsection{Results and Analysis}
We employed three state-of-the-art Multimodal Large Language Models (MLLMs) as our base frameworks: BLIP-2~\cite{li2023blip}, MiniGPT-4~\cite{zhu2023minigpt}, and Video-LLaMA~\cite{zhang2023video}. By integrating our proposed modules into these models, we developed the DynRsl-VLM. All models were fine-tuned under identical conditions to ensure a fair evaluation.

Our experimental outcomes are summarized in Table~\ref{tab:sota}, presenting the results on the NuInstruct test set. To streamline the presentation, we aggregated certain subtasks in our reporting; for instance, 'motion ego' and 'motion others' are combined, as are 'status ego' and 'status others'.

The results in Table~\ref{tab:sota} clearly indicate that integrating our modules leads to significant improvements in evaluation metrics across all tasks, confirming the effectiveness of DynRsl-VLM. Specifically, the incorporation of dynamic resolution processing enhances performance in perception risk assessment and reasoning tasks by approximately 4\% 3\% and 4\%, respectively. Moreover, DynRsl-VLM exhibits notable advancements in tasks sensitive to spatial distance and positional information, such as perception and prediction. This demonstrates that our approach effectively strengthens the model's ability to handle complex multimodal inputs.

\subsection{Ablation Study}

\definecolor{mygray}{gray}{.9}
\begin{table}
    \centering
    
    \begin{adjustbox}{max width=\linewidth}
    \begin{tabular}{l|c}
    \Xhline{1.5pt}
     \rowcolor{mygray}{\it \small \textbf{Method}}
    &{\small  \it \textbf{Accuracy}}
    \\

    \midrule
    Full Model &    34.8 \\
    \midrule
    w/o  DynRsl Image Extraction & 33.2  \\
    \midrule
    w/o DynRsl Image-Text Alignment & 34.1  \\
    \midrule
    w/o Both Modules & 31.1 \\
    \bottomrule
    \end{tabular}
    \end{adjustbox}
    \vspace{-3.0mm}
    \caption{\textbf{Impact of the DynRsl modules on model accuracy and efficiency.} The ablation study results show how the DynRsl Image Extraction and DynRsl Image-Text Alignment modules affect performance.
    }
    \label{tab:Ablation}
\end{table}
To assess the contributions of the proposed DynRsl Image Extraction and DynRsl Image-Text Alignment modules, we performed an ablation study by systematically removing each module from the full model. The results, summarized in Table~\ref{tab:Ablation}, demonstrate the impact of each component on the model's accuracy.

The Full Model, incorporating both modules, achieves the highest accuracy, confirming the effectiveness of our approach. When the DynRsl Image Extraction module is omitted (w/o Image Extraction Module), there is a noticeable decline in accuracy. This suggests that the Image Extraction module plays a crucial role in capturing relevant visual features. Similarly, removing the DynRsl Image-Text Alignment module (w/o Image-Text Alignment) also leads to reduced accuracy, indicating that this module is vital for effective multimodal alignment. The configuration without both modules (w/o Both Modules) shows the poorest performance in accuracy, highlighting the importance of each module individually and their synergistic effect when combined.

\section{Conclusion}
In this work, we explore vision-language modeling for autonomous driving tasks. We introduce DynRsl-VLM, which enhances Multimodal Large Language Models (MLLMs) by incorporating dynamic resolution processing across multiple subtasks. Our approach integrates two novel modules—the DynRsl Image Extraction and the DynRsl Image-Text Alignment—which together improve the handling of temporal, spatial, and resolution-specific details in complex multimodal inputs.

By avoiding feature loss and enhancing the model's ability to perceive the environment comprehensively, DynRsl-VLM improves the accuracy of autonomous driving tasks. Empirical evaluations on the NuInstruct dataset confirm the effectiveness of our method, highlighting its potential to advance multimodal understanding in autonomous driving scenarios.
{
    \small
    \bibliographystyle{ieeenat_fullname}

}


\end{document}